%% file: Main.tex
\documentclass[letterpaper, 10 pt, conference]{ieeeconf}

\IEEEoverridecommandlockouts
\usepackage{graphics} 
\usepackage{epsfig} 
\usepackage{times} 
\usepackage{amsmath} 
\usepackage{amssymb}  
\usepackage[table]{xcolor}
\usepackage[ruled,linesnumbered, noend]{algorithm2e}
\usepackage{longtable}

\usepackage{cite}
\usepackage{comment}
\usepackage{multirow}
\usepackage{graphicx}
\usepackage{wrapfig}
\usepackage{subfigure}
\usepackage{balance}
\usepackage{wrapfig}
\usepackage{booktabs}  
\usepackage{adjustbox}

\usepackage{soul, color}
\usepackage{colortbl}
\definecolor{Gray}{gray}{0.9}

\usepackage{pifont}
\newcommand{\ve}[1]{\mathbf{#1}} 
\newcommand{\tve}[1]{\tilde{\mathbf{#1}}} 

\usepackage{pifont}
\usepackage{graphicx} 

\title{Speech-to-Trajectory: Learning Human-Like Verbal Guidance \\for Robot Motion}

\author{Eran Beeri Bamani, Eden Nissinman, Rotem Atari, Nevo Heimann Saadon and Avishai Sintov 
\thanks{E. B. Bamani, E. Nissinman, R. Atari, N. Heimann Saadon and  A. Sintov are with the School of Mechanical Engineering, Tel-Aviv University, Israel.}
\thanks{This work was supported by the Israel Innovation Authority (grant No.
77857).}}

\begin{document}

\maketitle

\input{abstract}

\section{Introduction}
\label{sec:introduction}
\input{Introduction}
\section{Methods}

\input{Methods}
\label{sec:method}

\section{Model Evaluation}
\input{Evaluation}
\label{sec:Evaluation}

\section{Conclusions}
\input{Conclusions}

\bibliographystyle{IEEEtran}
\bibliography{ref}

\end{document}

%% file: Abstract.tex
\begin{abstract}
Full integration of robots into real-life applications necessitates their ability to interpret and execute natural language directives from untrained users. Given the inherent variability in human language, equivalent directives may be phrased differently, yet require consistent robot behavior. While Large Language Models (LLMs) have advanced language understanding, they often falter in handling user phrasing variability, rely on predefined commands, and exhibit unpredictable outputs. This letter introduces the Directive Language Model (DLM), a novel speech-to-trajectory framework that directly maps verbal commands to executable motion trajectories, bypassing predefined phrases. DLM utilizes Behavior Cloning (BC) on simulated demonstrations of human-guided robot motion. To enhance generalization, GPT-based semantic augmentation generates diverse paraphrases of training commands, labeled with the same motion trajectory. DLM further incorporates a diffusion policy-based trajectory generation for adaptive motion refinement and stochastic sampling. In contrast to LLM-based methods, DLM ensures consistent, predictable motion without extensive prompt engineering, facilitating real-time robotic guidance. As DLM learns from trajectory data, it is embodiment-agnostic, enabling deployment across diverse robotic platforms. Experimental results demonstrate DLM's improved command generalization, reduced dependence on structured phrasing, and achievement of human-like motion.

\end{abstract}

%% file: Introduction.tex
Natural and seamless communication between humans and robots is a critical challenge in robotics, particularly in converting high-level, often ambiguous verbal commands into desired robotic motion. The shift in the application domains of robotics, moving from predominantly repetitive tasks in controlled industrial settings to more varied and human-centric roles, underscores the growing importance of natural language interfaces. In domestic and healthcare environments, for example, users are often non-experts who lack the technical proficiency for traditional robot programming methods. The rise of Natural Language Processing (NLP) and Large Language Models (LLMs) has enabled more intuitive Human-Robot Interaction (HRI) \cite{Zhang2023}, yet existing models still face challenges in effectively translating verbal instructions into executable trajectories. 

Early methods primarily mapped voice inputs to pre-defined actions \cite{Roy2003,KressGazit2008}. While leveraging language structure, these approaches typically lacked learning capabilities and operated within fixed action spaces, thus limiting their domain applicability and hindering accessibility for novice users. The inherent ambiguity and phrasal variability of natural language pose a central challenge to translating human directives into robot-executable control commands \cite{Vandelaar2025}. This necessitates robust natural language understanding for robots, which involves parsing the linguistic structure of commands, identifying the intended actions and objects, and resolving any inherent ambiguities. Research in this area focuses on developing sophisticated techniques for semantic interpretation, empowering robots to extract the meaning and intent behind human language \cite{Liu2024}. This includes the ability to handle variations in phrasing, comprehend implicit instructions, and process incomplete or ungrammatical sentences.

\begin{figure}
    \centering
    \includegraphics[width=\linewidth]{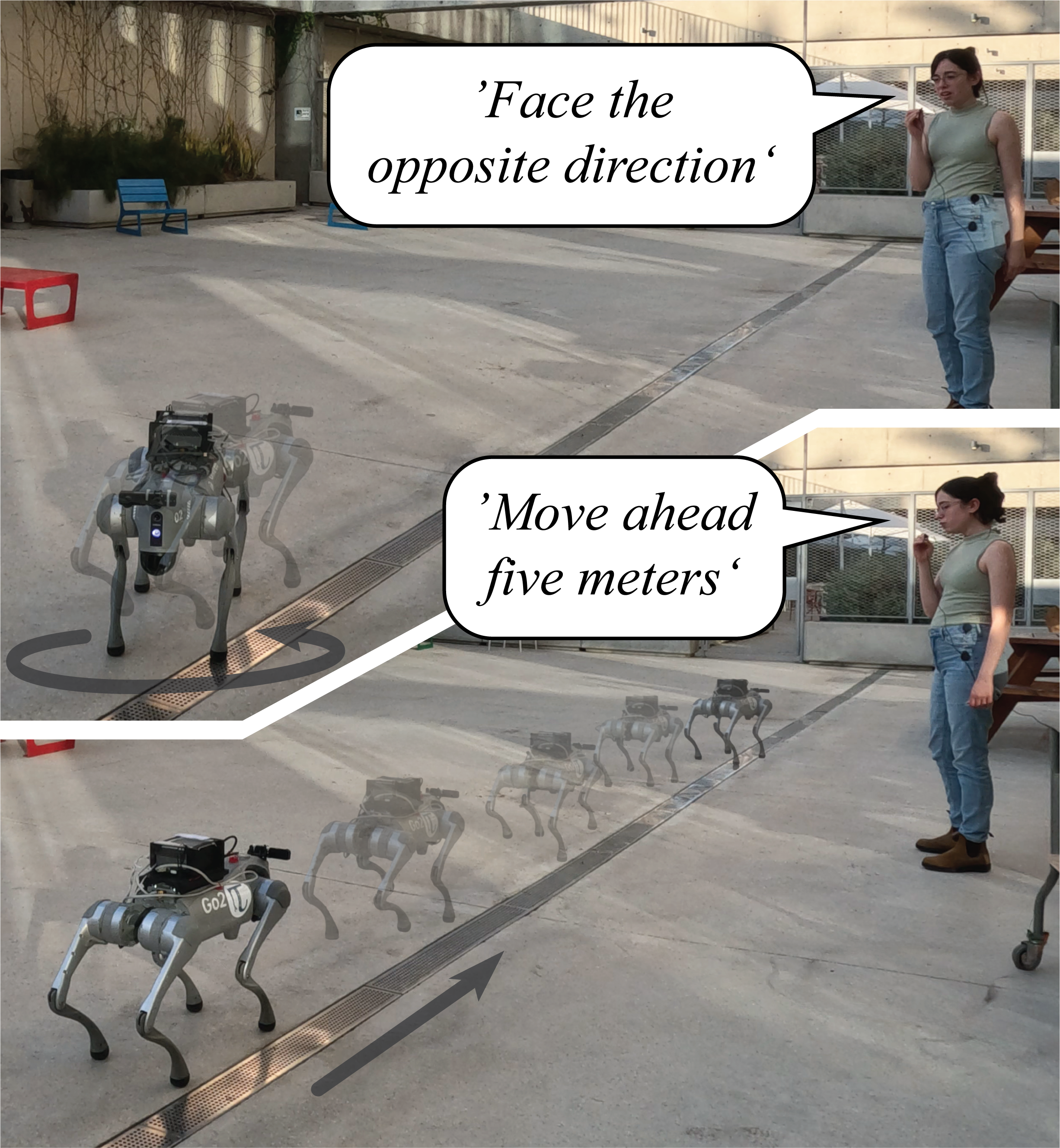}
    \caption{A user guides a quadruped robot to a target position using natural verbal commands. The proposed Directive Language Model (DLM) interprets these commands and translates them into motion trajectories. The same behavior can be triggered by differently phrased instructions, demonstrating the model’s ability to generalize across linguistic variations.}
    \label{fig:withGo1}
\end{figure}

Recent advancements in artificial intelligence, particularly in Deep Learning, has significantly advanced the field of robot guidance with natural language \cite{Su2023}. Deep learning models, especially those based on recurrent neural networks and the Transformer architecture, have greatly enhanced the ability to understand and process the complexities of natural language \cite{Torfi2020}. 
These have introduced new avenues for interpreting diverse human instructions without the need for explicitly defined reward signals \cite{radford2019better}. For instance, some approaches focus on directly mapping linguistic variations to actionable behaviors, which is particularly advantageous in scenarios where human feedback involves diverse expressions, corrections or preferences \cite{tellex2011understanding, stepputtis2020language}. However, a major gap remained in seamlessly integrating nuanced language processing with adaptive robotic behavior. The emergence of Large Language Models (LLMs), such as ChatGPT \cite{sanderson2023gpt}, represents a significant leap forward in the field. These models have demonstrated unprecedented language understanding and generation capabilities, opening up new possibilities for direct and intuitive control of robots using natural language.

Current research is actively exploring the use of LLMs for high-level task planning, enabling robots to understand complex goals expressed in natural language and to autonomously generate a sequence of actions required to achieve those goals \cite{zeng2023,Ding2023}. Some work combine Reinforcement Learning (RL) with LLM prompting, enabling agents to generalize better across diverse tasks and environments \cite{Sun2024}. 
However, these methods frequently depend on a predetermined output format, dictated by the prompt designer based on a specific robot's action space or policy code \cite{DeepMind2023,Li2023,wang2024prompt,Kannan2024}. Unlike traditional control models, LLMs generate stochastic responses, which can introduce variability in action selection and hinder reliability in critical tasks like robotic motion planning. This inconsistency makes it difficult to ensure repeatable and predictable behavior, especially in safety-sensitive applications. Also, extensive prompt engineering, required to align LLM outputs with desired actions, is time-consuming, requires expert crafting, lacks generalizability, and reduces system robustness \cite{Wang2025}. Finally, the computational demands of large-scale LLMs pose challenges for real-time deployment of robots.

A key research domain in directing robots to act upon natural language is often termed \textit{symbol grounding problem} – establishing a connection between the symbols used in natural language and the robot's sensory perceptions and interactions with the physical world \cite{kollar2013,jiang2023vima,Liu2024,Liu2024b}. However, these methods may not be well-suited for navigation tasks, as navigation involves a significantly larger configuration space compared to manipulation \cite{Boularias2015,Wen2023}. This increased complexity poses challenges for both training and inference in direct grounding approaches. To cope with such a problem, a motion planning layer is often added to bridge between the NLP and the action generation models \cite{hu2019}. However, these approaches depend on sensory perception and do not address the low-level actions that the robot must take. This approach may lead to suboptimal task performance and unexpected motions, resulting in user dissatisfaction. Additionally, they can affect performance expectations and lead to unnatural or potentially intimidating motions. In this work, on the other hand, we aim to achieve human-like performance that aligns with user expectations before incorporating sensory perception. Sensory inputs may later be used to impose motion constraints without dominating the motion itself.

\begin{figure*}
    \centering
    \includegraphics[width=\textwidth, keepaspectratio]{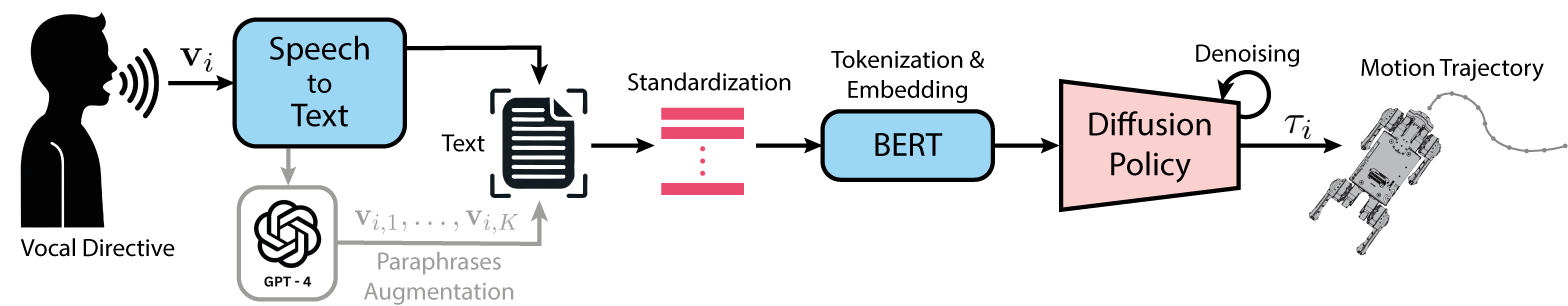} 
    \caption{Illustration of the proposed Directive Language Model (DLM) framework.}
    \label{fig:scheme}
\end{figure*}

While prior approaches try to learn high-level actions based on verbal commands and visual perception, in this paper, we address the learning of low-level action sequences determined solely based on verbal inputs (Figure \ref{fig:withGo1}). We introduce the Directive Language Model (DLM), a novel speech-to-trajectory framework designed for human-like verbal guidance of robots. The DLM framework is illustrated in Figure \ref{fig:scheme}. Unlike prior methods, DLM directly maps spoken command, without dependence on pre-defined or specific phrasing, to executable motion trajectories, enabling real-time robotic guidance. DLM follows a Behavior Cloning (BC) approach with data collected from multiple human participants who verbally guide or tele-operate virtual robots in a simulated environment. This allows the model to learn demonstrated motion patterns that align with human expectations and correspond to spoken commands. However, new participants may phrase commands differently than those seen during training. To address this, we employ GPT-based semantic augmentation, generating diverse paraphrases for the same trajectory, thereby improving generalization across varying speech patterns. Furthermore, the DLM framework leverages a diffusion policy-based trajectory generation framework, allowing for adaptive motion refinement and stochastic sampling to enhance trajectory flexibility. Because trajectory recording is conducted in simulation, DLM is embodiment-agnostic and can be deployed on any mobile robot.

Our key contributions are as follows:
\begin{itemize}
    \item We introduce the Directive Language Model (DLM), a novel speech-to-trajectory framework that translates natural spoken commands into executable low-level motion trajectories trajectories, enabling seamless and intuitive human-robot interaction.
    \item Unlike prior methods, DLM does not rely on specific pre-defined verbal structures but can generalize across varied linguistic expressions, improving usability for non-expert users.
    \item We use a dataset where human participants verbally guide or tele-operate virtual robots in a simulated environment, leveraging Behavior Cloning (BC) to learn motion patterns aligned with human expectations.
    \item We incorporate GPT-based data augmentation to enhance linguistic generalization, improving robustness to paraphrased or incomplete commands.  
    \item Since DLM learns from trajectory demonstrations rather than robot-specific control signals, it is applicable across different robotic platforms.
    \item Unlike LLM-based methods that require extensive prompt engineering and produce stochastic outputs, DLM ensures consistent, predictable behavior with lower computational demands.
    \item Experimental results demonstrate DLM's ability to accurately interpret both explicit and implicit commands, producing the corresponding expected trajectories.
\end{itemize}

%% file: Methods.tex
\subsection{Problem Formulation}
\label{sec:prob}

We aim for a policy with human-like interpretation that would enable semantic understanding and natural guidance of a robot. We consider the case of verbally guiding a robot to reach a desired $SE(2)$ pose in the environment. Let $\mathcal{V}$ be the space of all verbal motion directives that can be given to a robot to move on a planar space, represented as textual commands. Furthermore, a trajectory of the robot is defined by $\tau\in\mathcal{U}$ where $\mathcal{U}\subset SE(2)\times\ldots\times SE(2)$. The objective is to generate a robot trajectory, $\tau\in\mathcal{U}$, that accurately fulfills a given verbal command, $\ve{v}\in\mathcal{V}$. Hence, we search for a model $\Gamma:\mathcal{V}\to\mathcal{U}$ that maps a verbal command to a corresponding motion trajectory. We note that we focus on the generation of the trajectory and assume a trajectory-following control exists. 


\subsection{Data Collection}
\label{sec:data_collection}

To acquire model $\Gamma$, data is to be collected by labeling motion directives from $\mathcal{V}$ with corresponding motion trajectories from $\mathcal{U}$. Therefore, an interactive data collection framework is utilized, designed to guide robot motion in alignment with anticipated human behavior.

\begin{figure}
    \centering
    \includegraphics[width=\linewidth]{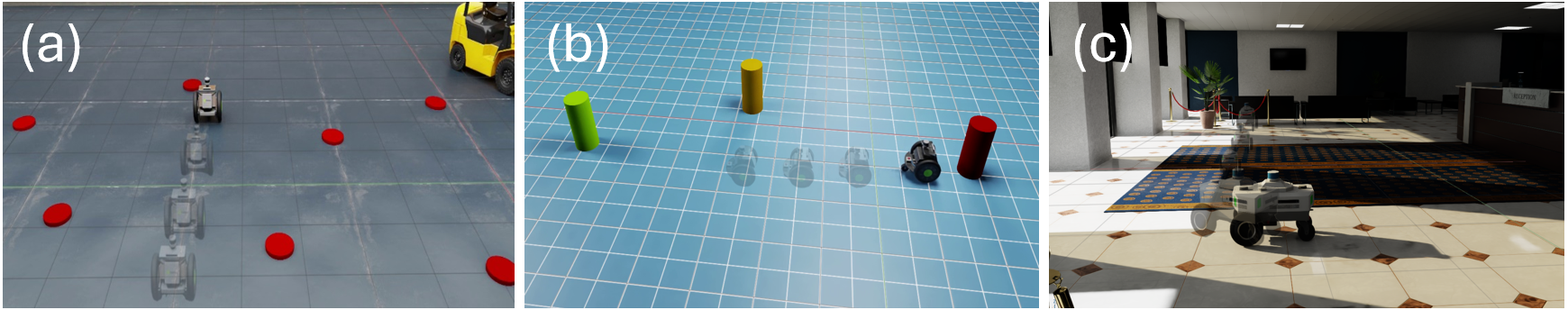} 
    \caption{Simulation environments in NVidia IsaacSim simulator. In these example scenarios, robot motion responses to commands (a) \texttt{Move forward four meters}, (b) \texttt{Go back three meters} and (c) \texttt{Go right five meters}, are demonstrated by a human driver.}
    \label{fig:data_collection}
\end{figure}

To enable extensive data collection without operating a real robot for a long period of time, and to enable a variety of environments, a simulated environment was created in the Nvidia IsaacSim simulator. The simulator is composed of a mobile ground robot moving on a flat surface with several obstacles, as demonstrated in Figure \ref{fig:data_collection}. Control of the robot is conducted with an Xbox controller, enabling a human driver to move it around in the simulated surface. Also, red markers were scattered across the simulated floor, each pinpointing a potential target for the robot to reach. On the other end, a leader participant is expected to provide vocal directives for the driver to obey. Any vocal command given by the leader is mapped to a textual format $\ve{v}\in\mathcal{V}$ using Whisper \cite{radford2022robust}. Whisper is a transformer-based model engineered for speech-to-text conversion. Its architecture is optimized for processing large volumes of weakly-supervised audio data, facilitating robust generalization across diverse acoustic environments and accents. In addition, we employ noise suppression pre-processing by combining classical spectral gating with the lightweight deep learning-based model RNNoise \cite{valin2018hybrid}, resulting in enhanced transcription accuracy of the Whisper model in noisy conditions. This approach enables robust transcription even in challenging acoustic environments, ensuring consistent and reliable performance of the speech-to-text system.

A collection session begins with the robot's random placement on the simulated floor. The leader then selects a red marker as a target to reach without revealing it to the driver. The leader directs the robot via free-form verbal commands to a microphone, improvising instructions throughout the iterative process without adhering to a fixed command vocabulary. Based solely on the leader's verbal command, the driver interprets and translates the instruction into robot movement using the controller, relying on their own subjective understanding. For example, when the leader says \textit{'Move forward five meters'}, the driver's interpretation and execution are influenced by their individual sense of distance. This subjective perspective, differing from the leader's, leads to variations in the robot's movement, similar to the behavior of humans in the same scenario. In each iteration, the command $\ve{v}_i\in\mathcal{V}$ is labeled with the driven trajectory $\tau_i\in\mathcal{U}$. This iterative process is repeated until the robot successfully reaches the leader's target. Across numerous sessions with different leaders and drivers, a dataset
\begin{equation}
\mathcal{D} = \{\ve{v}_i, \tau_i\}_{i=1}^N
\end{equation}
is acquired with $N$ labeled commands.






\begin{table*}[]
\caption{Paraphrase examples to three commands generated by GPT-4.}
\label{tb:command_variations}
\adjustbox{max width=\textwidth}{%
\begin{tabular}{lll}
\textbf{'Move forward 5 meters'}   & \textbf{'Move 4 meters to the left'}   & \textbf{'Turn slightly right'}  \\\hline
\texttt{'Advance 5 meters frontward'} & \texttt{'Advance 4 meters to the left side'} & \texttt{'Shift a smidge to the right'}  \\
\texttt{'Make a forward movement of 5 meters'} & \texttt{'Take a 4 meter step to the left'}  & \texttt{'Adjust your position a bit to the right'} \\
\texttt{'Move ahead 5 meters'} & \texttt{'Traverse 4 meters to the left'}  & \texttt{'Move a tiny bit right'} \\
\texttt{'Travel 5 meters in a forward path'} & \texttt{'Go 4 meters towards your left'}  & \texttt{'Go a little bit right'}   \\
\texttt{'Go forward a distance of 5 meters'} & \texttt{'Proceed 4 meters towards the left'}   & \texttt{'Bent a little to the right'} \\
\texttt{'Progress 5 meters straight ahead'}  & \texttt{'Head 4 meters in a leftward direction'}  & \texttt{'Move subtly right'}   \\
\texttt{'Head 5 meters straight onward'} & \texttt{'Make a leftward movement of 4 meters'}  & \texttt{'Make a slight rightward adjustment'} \\
\texttt{'Proceed forward 5 meters'} & \texttt{'Position yourself 4 meters to the left'} & \texttt{'Move slightly rightward'} \\
\texttt{'March 5 meters forward'} & \texttt{'Travel 4 meters to the left hand side'} & \texttt{'Shift a small amount to the right'}    \\
\end{tabular}}
\end{table*}


\subsection{Semantic Augmentation}
\label{sec:augmentation}

We aim to enhance the robot's semantic understanding and responsiveness to diverse commands. Despite the inclusion of multiple leaders with varied linguistic styles, dataset $\mathcal{D}$ remains insufficient to represent the full distribution of the command vocabulary in $\mathcal{V}$. To further improve robustness against linguistic variability, we employ a paraphrase generation approach. For each command $\ve{v}_i\in\mathcal{D}$, we generate $K$ paraphrased variations by prompting GPT-4 (e.g., \texttt{'Generate <K> variations of the following command: <command>'}). Then, the generated paraphrases $\{\ve{v}_{i,1},\ldots,\ve{v}_{i,K}\}$ are labeled with the same trajectory $\tau_i$. The original and generated commands are added to a new dataset of the form
\begin{equation}
\mathcal{H} = \{\ve{v}_j, \tau_j\}_{j=1}^{N(K+1) }.
\end{equation}
These paraphrases augment the training dataset, enabling the semantic parser to learn from a richer set of expressions and thereby generalize better to unseen commands. Examples of simple commands and their generated paraphrases are given in Table \ref{tb:command_variations}. This augmentation strategy ensures the system can recognize and accurately interpret semantically equivalent instructions expressed differently, leading to more robust HRI.



\subsection{Directive Language Model (DLM)}

The proposed DLM, illustrated in Figure \ref{fig:scheme}, implements the $\Gamma$ model, mapping a verbal command $\ve{v}_i$ into a motion trajectory $\tau_i$. As mentioned above, any verbal command is mapped to a textual representation using a pre-trained speech-to-text model. Subsequently, to effectively manage this linguistic diversity, we introduce a separate textual standardization step. In this step, the paraphrased textual commands in $\mathcal{H}$ are transformed into semantically consistent textual forms, thus reducing semantic ambiguity \cite{ide1998introduction,bubeck2023sparks,yoo2021gpt3mix}. Following standardization, each extracted command undergoes tokenization and embedding using Bidirectional Encoder Representations from Transformers (BERT) \cite{Acheampong2021} to yield a unified semantic representation. Unlike autoregressive models like GPT, which predict subsequent tokens in a sequence, BERT leverages bidirectional embeddings, analyzing both preceding and succeeding words to generate contextually rich representations. This bidirectionality is crucial for robotic navigation, where understanding the full semantic context ensures accurate interpretation of user intent. Paraphrases with similar meanings are processed into a shared embedding space. 

Given a standardized input sentence, we first apply tokenization, yielding a sequence of tokens $T = \{t_1, t_2, \dots, t_m\}$, where each token $t_j$ represents either a word or sub-word unit. These tokens are then mapped into embedding vectors $\ve{E} = \{\ve{e}_1, \ve{e}_2, \dots, \ve{e}_m\}$ using the BERT embedding function, where $\mathbf{e}_j \in \mathbb{R}^d$ for some embedding dimension $d$. While different paraphrases produce distinct token sequences, the embedding process positions them closely within the semantic space, ensuring a standardized input for downstream trajectory generation. By enforcing semantic consistency within the speech-to-text processing pipeline, achieved through BERT embeddings, the system improves its adaptability to diverse linguistic styles and user preferences. This adaptability empowers the robot to accurately interpret nuanced commands, even in real time, significantly enhancing human-robot interaction and facilitating robust, reliable trajectory generation.

Given the embedding $\ve{E}_j$ representing command $\ve{v}_j$, our method utilizes a Diffusion Policy (DP) to model the conditional distribution $p(\tau_j | \ve{E}_j)$ \cite{Chi2023diffusion}. This approach, building upon the generative framework of DP, differs from prior work that relies on visual inputs. By conditioning on textual embeddings, we enable direct text-based control, generating multiple feasible motion plans through stochastic sampling. The DP approximates the distribution by sampling an action trajectory $\tilde{\tau}_{j}^{(0)}=\{\ve{x}_{j,1}^{(0)},\ldots,\ve{x}_{j,h}^{(0)}\}$ with length $h$ where $\ve{x}_{j,i}^{(k)}\in SE(2)$ is a point along $\tilde{\tau}_{j}^{(k)}$ at denoising iteration $k$. Point $\ve{x}_{j,1}^{(0)}$ is sampled from a Gaussian distribution. Subsequently, $K$ denoising iterations are performed, progressively refining the action trajectories to produce a noise-free output $\tilde{\tau}_{j}^{(K)}$. This denoising process follows the iterative formula  
\begin{equation}
    \tilde{\tau}_{j}^{(k+1)} = \alpha \cdot \left(\tilde{\tau}_{j}^{(k)} - \gamma \epsilon_\theta(\ve{E}_j, \tilde{\tau}_{j}^{(k)}, k)\right) + \mathcal{N}(0, \sigma^2 I),
\end{equation}
where noise schedule functions $\alpha$, $\gamma$ and $\sigma$ control the learning rate during denoising. The function $\epsilon_\theta$ denotes a noise prediction network, parameterized by $\theta$, tasked with reconstructing trajectories in accordance with the inherent constraints of robotic motion as demonstrated in the collected data. The noise prediction network model is implemented using a transformer-based architecture, which receives both the embedded semantic condition $\ve{E}_j$ and a noise-corrupted trajectory target $\tilde{\tau}_j^{(k)}$. Its output is a noise vector representing the estimated perturbation applied to the trajectory, which is progressively removed through iterative denoising steps to recover an accurate trajectory aligned with the verbal instruction. 

In the original DP approach, a masking mechanism used visual input to determine trajectory length $h$. Since our method relies solely on textual commands without explicit environmental data, we introduce an Adaptive Trajectory Length Determination (ATLD) mechanism. Instead of a predefined masking step, we employ a decision-making component that dynamically determines when to end the generated trajectory. Let $\tve{\tau}_j=\{\ve{x}_{j,1},\ldots,\ve{x}_{j,H}\}$ represent a generated trajectory. If the number of generated points $H$ is redundant with respect to demonstrated trajectories, DP will misplace advanced trajectory points. Hence, we define a termination criterion based on trajectory smoothness in which point $1<h\leq H$ is the one that satisfies 
\begin{equation}
\|\ve{x}_{j,i} - \mathbf{x}_{j,i-\lambda}\| < \epsilon,~\forall i=1,\ldots,h+\lambda-1
\end{equation}  
where $\lambda$ is a window length capturing recent dynamics, and $\epsilon$ is a threshold for minimal displacement, indicating goal convergence. By analyzing trajectory behavior, the model autonomously infers when motion sufficiently aligns with demonstrated user intent, ensuring flexible and precise trajectories without predefined lengths.

%% file: Evaluation.tex
In this section, we evaluate the proposed DLM's ability to convert natural language directives into desired robotic motion. Our evaluations include state-of-the-art comparison, ablation study, robustness analysis and guiding experiments with a real robot. 
The data collection and experiments were conducted with the approval of the 
ethics committee at Tel-Aviv University under application No. 0010028. 
All computations were accelerated using four NVIDIA GeForce RTX 3080TI GPUs with 16GB RAM each. Videos demonstrating the robot experiments, both in simulation and real-world environments, are available in the supplementary material.


\subsection{Dataset \& Training}

Data was collected as described in Section \ref{sec:data_collection}. The process involved 14 and 19 different leaders and drivers, respectively. During data collection, leaders and drivers had no communication beyond a single directive per iteration. Upon receiving a command, drivers guided the virtual robot in the simulated environment based on their subjective understanding, without knowledge of the session's final target. Across 680 sessions, a dataset $\mathcal{D}$ of $N=18,500$ labeled samples were collected. In the semantic augmentation of Section \ref{sec:augmentation}, we generated $K=30$ paraphrases for each $\ve{v}_j\in\mathcal{D}$ using GPT-4, yielding augmented dataset $\mathcal{H}$ with 541,814 labeled samples. In addition to dataset $\mathcal{H}$ used for training, we collected an independent test set comprising 6,590 labeled samples collected with 10 different leaders and drivers.

The DLM was trained using the Adam optimizer with optimized hyperparameters: a batch size of 64, a linearly increasing learning rate from 0.0001 to 0.002, and a weight decay factor of $1.25\times10^{-6}$. The masking mechanism was conducted with $\lambda=7$, $\epsilon=0.03$ and $H=22$. Training was conducted over 30 epochs, with optimization performed on learning rate schedules and weight regularization. The loss function, designed to guide the model towards generating smooth and precise motion trajectories, combined the Root Mean Squared Error (RMSE) for positional accuracy and the Mean Absolute Orientation Error (MAOE) for angular accuracy. These metrics were jointly used to supervise both noise prediction and length outputs. 

\begin{table}
    \centering
    \caption{Comparative analysis for various models in speech-to-trajectory tasks}
    \label{tb:ComparisonModels}
    \begin{tabular}{lcccc}
        \toprule
        \multirow{2}{*}{Model} & SR & RMSE  & MAOE  & Inference \\
        & (\%) & (cm) & ($^\circ$) & time (ms) \\
        \midrule
        Wav2Vec & 83.2 $\pm$ 2.5 & 45.0 $\pm$ 6.4 & 9.1 $\pm$ 1.2 & 73 \\
        PaLM-E & 85.7 $\pm$ 2.7 & 37.0 $\pm$ 2.3 & 7.9 $\pm$ 0.9 & 138 \\
        T5 & 71.5 $\pm$ 3.8 & 105.0 $\pm$ 6.6 & 14.0 $\pm$ 2.5 & \cellcolor[HTML]{C0C0C0}44 \\
        D3QN & 78.4 $\pm$ 4.1 & 89.0 $\pm$ 5.0 & 9.7 $\pm$ 1.1 & 81 \\
        VIMA & 84.9 $\pm$ 2.8 & 42.0 $\pm$ 2.5 & 8.2 $\pm$ 1.3 & 85 \\
        DLM & \cellcolor[HTML]{C0C0C0}95.6 $\pm$ 1.2 & \cellcolor[HTML]{C0C0C0}9.0 $\pm$ 3.0 & \cellcolor[HTML]{C0C0C0}2.8 $\pm$ 0.6 & 88 \\
        \bottomrule
    \end{tabular}%
\end{table}


\subsection{Model evaluation}

Using dataset $\mathcal{H}$, we conduct a comparative evaluation of the proposed DLM against five state-of-the-art models, each adapted for our speech-to-trajectory task. Wav2Vec \cite{Baevski2020} is a self-supervised speech representation model that processes raw audio inputs, fine-tuned with a regression head to predict motion trajectories directly from spoken commands. PaLM-E \cite{Driess2023}, a multimodal language model designed for embodied AI, was adapted to process transcribed commands and trained with a custom regression decoder to output trajectory coordinates. T5 \cite{Raffel2020}, a transformer-based text-to-text model, was modified to generate motion trajectories as a sequence prediction task, tokenizing each pose for structured output. D3QN \cite{Zhou2019} is a reinforcement learning model using a double deep Q-network with prioritized experience replay. It was adapted by discretizing the action space and conditioning decisions on semantic embeddings from a pretrained BERT model. Trained via imitation learning on expert trajectories, it sequentially selected discrete actions to reconstruct a continuous path from verbal commands. VIMA \cite{jiang2023vima}, originally designed for vision-language robotic manipulation, was adapted by removing its visual inputs and using only textual prompts to generate motion plans through its transformer-based policy. These models represent a diverse set of approaches, encompassing speech processing, multimodal reasoning, text generation, reinforcement learning, and vision-language understanding, providing a robust baseline for evaluating the DLM’s performance.

Comparison between the trained models is evaluated over the test set with four key metrics: target reach Success Rate (SR) within a 10 cm position error, RMSE, MAOE and inference time. 
RMSE and MAOE evaluate the mean position and orientation errors, respectively, between demonstrated trajectories and generated ones, based on input commands.
Table \ref{tb:ComparisonModels} presents the comparative performance metrics for all evaluated models. The results demonstrate that DLM achieves superior trajectory generation accuracy, as evidenced by higher SR values and significantly lower RMSE and MAOE, compared to all other models. While DLM does not exhibit the lowest inference time, it remains capable of facilitating real-time performance.

\begin{figure}
    \centering
    \includegraphics[width=\linewidth]{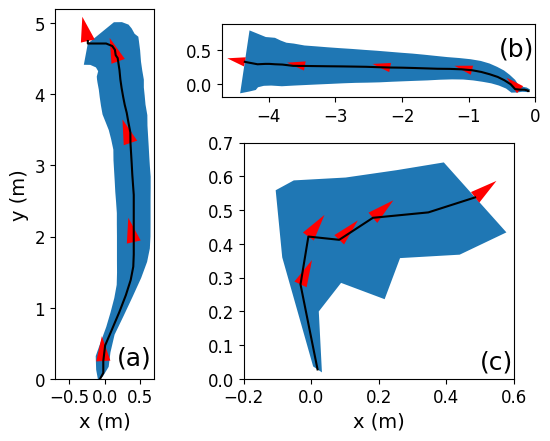} 
    \caption{Mean (black curve) and standard deviations (blue) of outputted trajectories to paraphrases of the commands (a) \texttt{'Move forward 5 meters'}, (b) \texttt{'Move 4 meters to the left'} and (c) \texttt{'Turn slightly right'}, given in Table \ref{tb:command_variations}. The red arrows indicate the means of the required robot orientations along the paths.}
    \label{fig:Paths}
\end{figure}

We further assess the DLM's consistency in generating trajectories for different paraphrases of the same command. Using the example commands and paraphrases in Table \ref{tb:command_variations}, we analyze the distribution of output trajectories in Figure \ref{fig:Paths}. The results demonstrate a high degree of accuracy in the generated motions, aligning closely with the desired directives, and exhibiting low variance in the distributions. Consequently, the DLM demonstrates a robust consistency to paraphrase variations.

\subsection{Ablation Study} 

To validate the individual contributions of various components within the DLM, an ablation study was conducted. The components subjected to analysis included the GPT-based data augmentation, data standardization, and the BERT embedding and tokenization layer. The findings of this study are reported in Table \ref{tb:AblationResults}. The results demonstrate a distinct contribution from each component towards the reduction of both RMSE and MAOE, and an increase in target reaching success rate. Notably, the findings underscore the significant role of data augmentation in diversifying the command space and enhancing the model's comprehension.

\begin{table}
    \centering
    \caption{Ablation study for the DLM components}
    \label{tb:AblationResults}
    \begin{tabular}{lccc}
        \toprule
        Model Variant & SR (\%) & RMSE (cm) & MAOE (°) \\
        \midrule
        w/o Standardization & 91.5 $\pm$ 1.4 & 22.5 $\pm$ 4.5 & 5.8 $\pm$ 0.8 \\
        w/o ATLD & 88.2 $\pm$ 2.3 & 23.1 $\pm$ 3.8 & 6.1 $\pm$ 0.4 \\
        w/o BERT & 83.6 $\pm$ 2.3 & 31.0 $\pm$ 7.0 & 7.9 $\pm$ 1.2 \\
        w/o GPT-4 Augmentation & 81.4 $\pm$ 2.5 & 35.0 $\pm$ 8.0 & 8.7 $\pm$ 1.4 \\
        Full DLM & \cellcolor[HTML]{C0C0C0}95.6 $\pm$ 1.0 & \cellcolor[HTML]{C0C0C0}9.0 $\pm$ 3.0 & \cellcolor[HTML]{C0C0C0}2.8 $\pm$ 0.6 \\
        \bottomrule
    \end{tabular}%
\end{table}


\subsection{Robustness to Noisy and Incomplete Commands} 

In real-world environments, voice commands are often distorted by noise, microphone limitations, or overlapping speech. These distortions can result in missing words, partial commands, or unintended word insertions, leading to ambiguous instructions. For instance, signal loss may truncate commands (e.g., \texttt{'Move forward five'} instead of \texttt{'Move forward five meters'}), or background noise may omit key context (e.g., \texttt{'Turn left'} instead of \texttt{'Turn left at the door'}). External speech interference can also alter meanings, such as capturing \texttt{'Go forward'} when another person says \texttt{'right'} in the background.

To assess the model’s robustness, we evaluate its performance under three corruption scenarios: word dropout, where a keyword is randomly omitted; sentence truncation, where only partial commands are received; and mixed-speaker input, where irrelevant words are injected. For each type, we corrupted and tested 925 commands in the test set. These experiments quantify the model’s ability to reconstruct and interpret corrupted speech. Table \ref{tb:command_corruption} presents SR, RMSE and MAOE for each corruption type.

The model effectively handles minor word omissions but struggles with mixed-speaker interference, underscoring the importance of contextual embeddings and sequence reconstruction. While it compensates well for word dropouts, severe truncation or mixed-speech interference significantly reduces accuracy. Future work could integrate self-supervised learning for sentence reconstruction and filtering mechanisms to mitigate external speech contamination.

\begin{table}
    \centering
    \caption{Performance under corrupted commands}
    \label{tb:command_corruption}
    \begin{tabular}{lccc}
        \toprule
        Corruption type & SR (\%) & RMSE (cm) & MAOE ($^\circ$) \\
        \midrule
        Single word dropout         & 90.2 & 18.4 & 5.1 \\
        Sentence truncation         & 76.3 & 34.7 & 7.8 \\
        Mixed-speaker input         & 61.5 & 58.3 & 13.4 \\
        \bottomrule
    \end{tabular}
    \footnotesize
\end{table}


\subsection{Natural Guidance Evaluation in Simulation} 

In the next experiment, we evaluate the DLM's ability to support a human leader in naturally guiding a robot, compared to directing a human expert driver. For this purpose, we recruited ten random participant with no prior experience in robotics or familiarity with our research. The participants were tasked with verbally guiding both the model-driven system and an expert human driver toward a predefined goal. Each participant encountered a different simulated scenario with varied environmental obstacles and initial pose of the robot. The participant was instructed to select a target from the floor markings while keeping it undisclosed. To maintain fairness, neither the expert driver nor the model had prior knowledge of the target location before execution. Furthermore, aside from the participant's iterative commands, no additional communication was permitted between the participant and the human driver. The goal was considered reached when the robot entered a one-meter radius around the target. This setup ensured that navigation relied solely on the effectiveness of the participant's guidance.

Evaluation metrics include final positional error, number of control steps with commands, total session time from start to reaching the goal, and a subjective rating reflecting the participant’s perception of robot effective compliance. Participants independently provided subjective ratings, ranging from 1 to 100, to both the expert driver and the DLM, based on their perceived compliance with the given directives. Table \ref{tb:random_guidance_experiment} presents the average results for ten expert- and model-driven sessions. All sessions concluded with reaching the goal. The results show that while the expert driver consistently demonstrated good directive compliance and efficient task completion, the model-driven approach achieved comparable navigation performance in terms of the number of steps and motion time, with slightly decreased accuracy. While expert drivers received slightly higher subjective scores than the DLM, the model's ratings remained high and comparable. Notably, lower scores for the DLM often correlated with greater trajectory deviations, highlighting the need to refine trajectory generation to better match human expectations. However, potential bias may have influenced scoring, as participants were aware of whether the driver was human. Although this study lacked the resources to control for this factor, future work could mitigate bias by concealing the human driver, ensuring participants perceive all sessions as autonomous.


\begin{table}
    \centering
    \caption{Comparison of expert- vs. model-driven robot guidance in simulation}
    \label{tb:random_guidance_experiment}
    \begin{tabular}{lcccc}
        \toprule
        Driver & Error (m) & Num. Steps & Time (s) & Subj. rating \\
        \midrule
        Expert & 0.41 $\pm$ 0.14 & 6.1 $\pm$ 1.58 & 45.7 $\pm$ 24.3 & 86.5 $\pm$ 10.9 \\
        DLM    & 0.74 $\pm$ 0.19 & 6.1 $\pm$ 2.38 & 43.7 $\pm$ 27.7  & 78 $\pm$ 10.3  \\
        \bottomrule
    \end{tabular}
\end{table}


\subsection{Guiding a quadruped} 

We further assess the DLM’s performance in real-world robot guidance using the Unitree Go2 quadruped. In this setup, a base computer ran the DLM, processing voice commands from a connected microphone and controlling the robot’s motion based on the generated trajectory. Due to wireless communication difficulties, the robot was operated using an Ethernet tether. Experiments were conducted in an open space of approximately 64~$m^2$ area, over several navigation scenarios. The scenarios include: reaching a single target with a clear path (without obstacles); reaching two targets sequentially with a clear path; reaching a single target with one obstacle between the robot and the target; reaching a single target with three obstacle scattered between the robot and the target; and reaching a single target while receiving implicit directives. In all scenarios, participants naturally provided verbal commands as they wished, without any constraints or instructions on wording. In the implicit directive scenario, we tested the DLM's ability in infer about the desired trajectory without an explicit command. For example, the participants may say \texttt{'I am standing on your left with a distance of three meters'}, expecting the robot to move to their location. Each scenario was tested across three sessions with different participants, and the results were averaged. In each session, the robot and target were randomly positioned within the open space to ensure variability.
\begin{table}
    \centering
    \caption{Real-Time Experiment Results}
    \label{tab:real_time_experiments}
    \begin{tabular}{lccc}
        \toprule
        Scenario & Error (m) & Num. Steps & Time (s) \\
        \midrule
        Single target w/ clear path & 0.23 & 3.6 & 52  \\
        Two targets w/ clear path & 0.25 & 7 & 103 \\
        Single target w/ obstacle  & 0.4 & 2.8 & 72 \\
        Single target w/ three obstacles & 0.2 & 3 & 93 \\
        Single target w/ implicit directives & 0.43 & 1.33 & 53 \\
        \bottomrule
    \end{tabular}
\end{table}

\begin{figure}
    \centering
    \includegraphics[width=\linewidth]{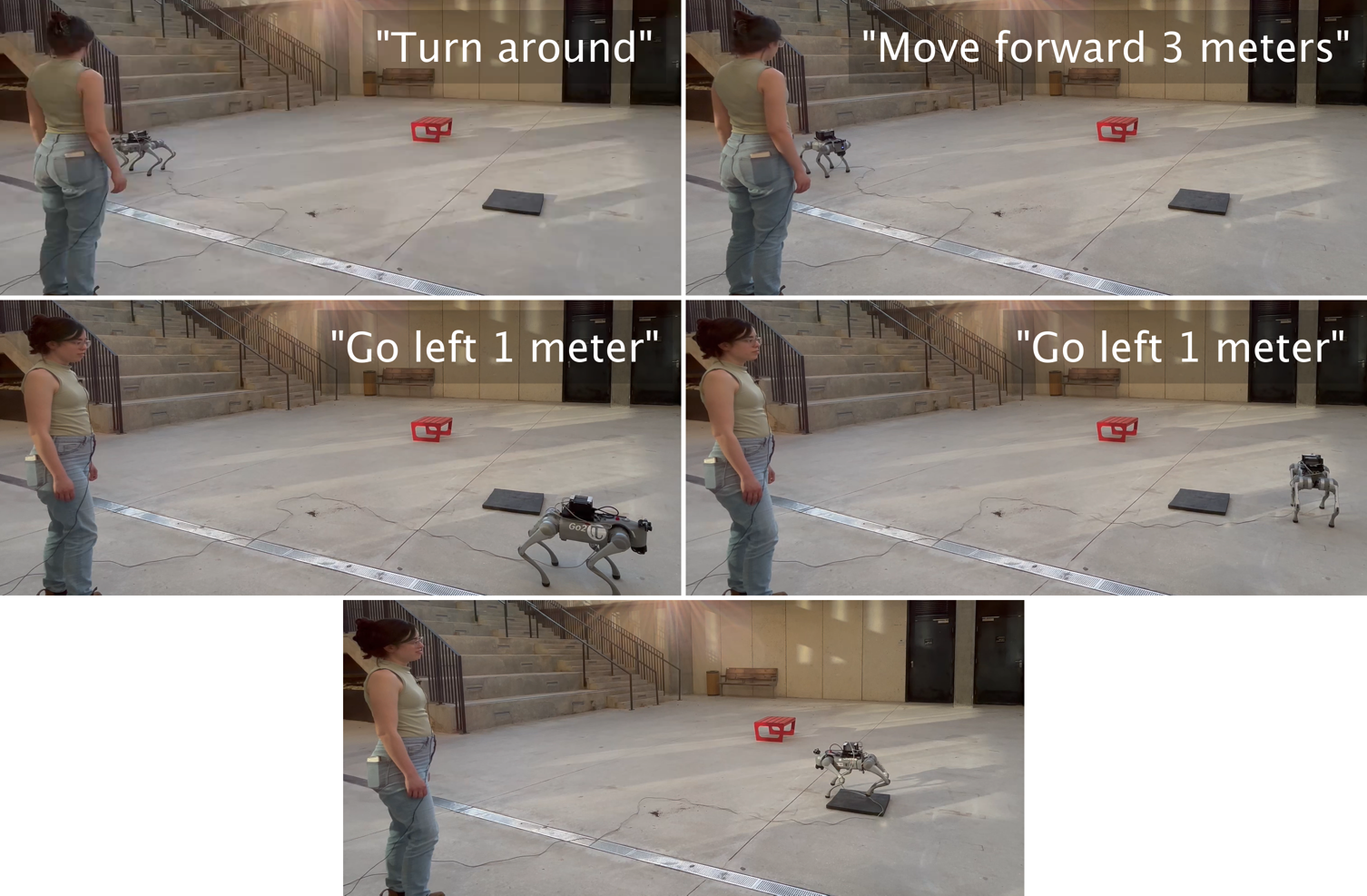} 
    \caption{Sequential robot navigation towards a target square, achieved through four iterations of verbal command input processed by the DLM.}
    \label{fig:exp_4_steps}
\end{figure}

\begin{figure}
    \centering
    \includegraphics[width=\linewidth]{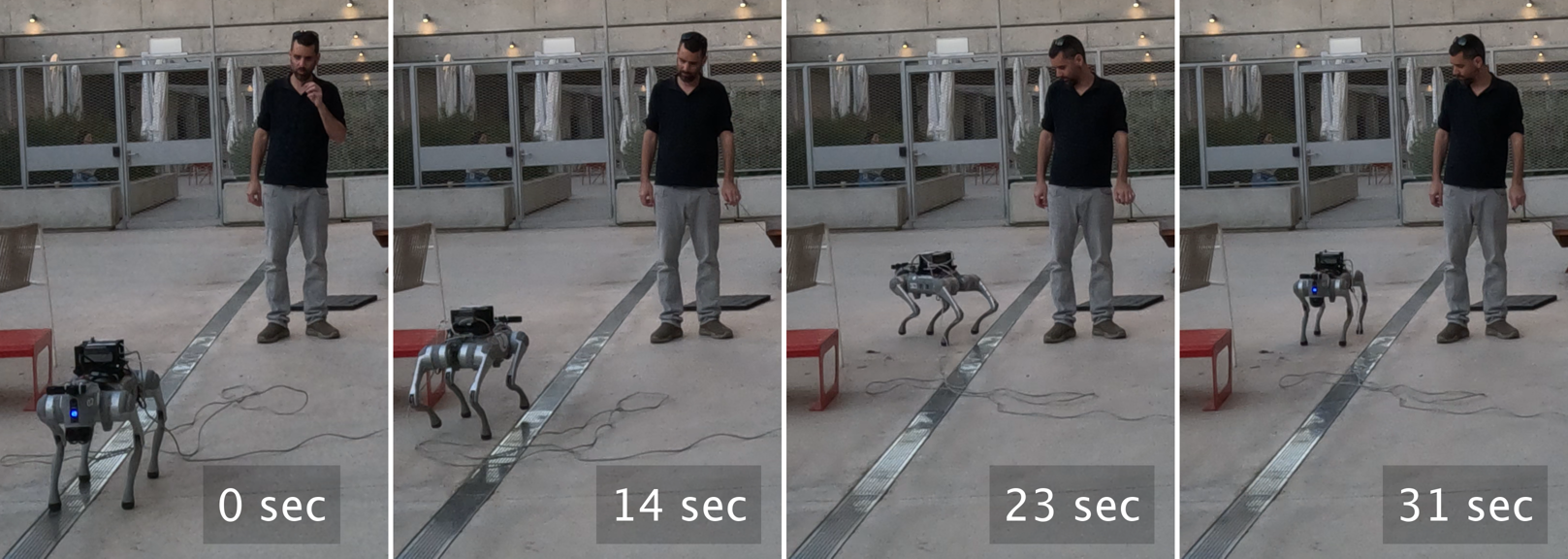} 
    \caption{Robot's motion response to the implicit directive: \texttt{'I am standing behind you 3 meters'}, demonstrating navigation towards the user without an explicit command. }
    \label{fig:exp_standing}
\end{figure}

Table \ref{tab:real_time_experiments} presents the mean error from the center of the robot to the target, mean number of directives steps required to reach the target, and the mean total gross time of the session from the first directive to reaching the target. The results show that all sessions were concluded with reaching the target, exhibiting a low positioning error. Furthermore, the number of directive steps and completion time are low and correspond to the complexity of each scenario. Across all scenarios, and especially in the implicit directives case, the robot effectively extracted motion instructions by filtering out non-essential terms and focusing on key phrases. Figures \ref{fig:withGo1} and \ref{fig:exp_4_steps} show snapshots of different scenarios with one target. Figure \ref{fig:exp_standing} shows a successful trial where the user provided an implicit directive to the robot by merely stating the position relative to the robot. The results highlight the ability of DLM to understand natural directives and move as humans expect.


%% file: Conclusions.tex
In this letter, we have addressed the problem of intuitive and natural low-level guidance of a mobile robot using verbal commands. We proposed the DLM framework for speech-to-trajectory mapping, trained using data collected from human demonstrations. DLM leverages BC and GPT-based semantic augmentation to improve linguistic generalization and adaptability. Unlike previous approaches that rely on predefined command structures, our method enables flexible and intuitive HRI without requiring extensive prompt engineering. Through simulation-based training and diffusion policy-based trajectory generation, we achieved human-like motion execution that aligns with user expectations. Our results in simulation and on a real robot highlight the effectiveness of speech-driven control in enhancing natural interaction with robotic platforms.

Given our approach's low-level motion control, future research will explore the integration of visual perception for context-aware command compliance and human gesture recognition. This would enable users to incorporate high-level task information alongside spatial directives. Furthermore, RL integration may enhance adaptability in dynamic environments and facilitate efficient real-time decision-making. Additionally, closed-loop verbal feedback could be implemented to rectify faulty trajectories in real-time.

%% file: Main.bbl
\begin{thebibliography}{10}
\providecommand{\url}[1]{#1}
\csname url@samestyle\endcsname
\providecommand{\newblock}{\relax}
\providecommand{\bibinfo}[2]{#2}
\providecommand{\BIBentrySTDinterwordspacing}{\spaceskip=0pt\relax}
\providecommand{\BIBentryALTinterwordstretchfactor}{4}
\providecommand{\BIBentryALTinterwordspacing}{\spaceskip=\fontdimen2\font plus
\BIBentryALTinterwordstretchfactor\fontdimen3\font minus \fontdimen4\font\relax}
\providecommand{\BIBforeignlanguage}[2]{{%
\expandafter\ifx\csname l@#1\endcsname\relax
\typeout{** WARNING: IEEEtran.bst: No hyphenation pattern has been}%
\typeout{** loaded for the language `#1'. Using the pattern for}%
\typeout{** the default language instead.}%
\else
\language=\csname l@#1\endcsname
\fi
#2}}
\providecommand{\BIBdecl}{\relax}
\BIBdecl

\bibitem{Zhang2023}
C.~Zhang, J.~Chen, J.~Li, Y.~Peng, and Z.~Mao, ``Large language models for human–robot interaction: A review,'' \emph{Biomimetic Intelligence and Robotics}, vol.~3, no.~4, p. 100131, 2023.

\bibitem{Roy2003}
D.~Roy, K.-Y. Hsiao, and N.~Mavridis, ``Conversational robots: Building blocks for grounding word meaning,'' in \emph{Workshop on Learning Word Meaning from Non-Linguistic Data}, 2003, pp. 70--77.

\bibitem{KressGazit2008}
H.~Kress-Gazit, G.~E. Fainekos, and G.~J.~P. and, ``Translating structured english to robot controllers,'' \emph{Advanced Robotics}, vol.~22, no.~12, pp. 1343--1359, 2008.

\bibitem{Vandelaar2025}
T.~van~de Laar, Z.~Zhang, S.~Qi, S.~Haesaert, and Z.~Sun, ``{VernaCopter}: Disambiguated natural-language-driven robot via formal specifications,'' \emph{arxiv 2409.09536}, 2025.

\bibitem{Liu2024}
R.~Liu, Y.~Guo, R.~Jin, and X.~Zhang, ``A review of natural-language-instructed robot execution systems,'' \emph{AI}, vol.~5, pp. 948--989, 2024.

\bibitem{Su2023}
H.~Su, W.~Qi, J.~Chen, C.~Yang, J.~Sandoval, and M.~A. Laribi, ``Recent advancements in multimodal human–robot interaction,'' \emph{Frontiers in Neurorobotics}, vol.~17, 2023.

\bibitem{Torfi2020}
A.~Torfi, R.~A. Shirvani, Y.~Keneshloo, N.~Tavvaf, and E.~A. Fox, ``Natural language processing advancements by deep learning: A survey,'' \emph{ArXiv}, vol. abs/2003.01200, 2020.

\bibitem{radford2019better}
A.~Radford, J.~Wu, D.~Amodei, D.~Amodei, J.~Clark, M.~Brundage, and I.~Sutskever, ``Better language models and their implications,'' \emph{OpenAI blog}, vol.~1, no.~2, 2019.

\bibitem{tellex2011understanding}
S.~Tellex, T.~Kollar, S.~Dickerson, M.~Walter, A.~Banerjee, S.~Teller, and N.~Roy, ``Understanding natural language commands for robotic navigation and mobile manipulation,'' in \emph{AAAI Conference on Artificial Intelligence}, vol.~25, no.~1, 2011, pp. 1507--1514.

\bibitem{stepputtis2020language}
S.~Stepputtis, J.~Campbell, M.~Phielipp, S.~Lee, C.~Baral, and H.~Ben~Amor, ``Language-conditioned imitation learning for robot manipulation tasks,'' \emph{Advances in Neural Information Processing Systems}, vol.~33, pp. 13\,139--13\,150, 2020.

\bibitem{sanderson2023gpt}
K.~Sanderson, ``{GPT-4} is here: what scientists think,'' \emph{Nature}, vol. 615, no. 7954, p. 773, 2023.

\bibitem{zeng2023}
F.~Zeng, W.~Gan, Y.~Wang, N.~Liu, and P.~S. Yu, ``Large language models for robotics: A survey,'' \emph{arXiv2311.07226}, 2023.

\bibitem{Ding2023}
Y.~Ding, X.~Zhang, C.~Paxton, and S.~Zhang, ``Task and motion planning with large language models for object rearrangement,'' in \emph{IEEE/RSJ Int. Conf. on Intel. Rob. and Sys.}, 2023, pp. 2086--2092.

\bibitem{Sun2024}
J.~Sun, Q.~Zhang, Y.~Duan, X.~Jiang, C.~Cheng, and R.~Xu, ``Prompt, plan, perform: {LLM}-based humanoid control via quantized imitation learning,'' in \emph{IEEE Int. Conf. on Robotics and Automation}, 2024.

\bibitem{DeepMind2023}
G.~DeepMind, ``Demonstrating large language models on robots,'' in \emph{Robotics: Science and Systems}, 2023.

\bibitem{Li2023}
G.~Li, X.~Han, P.~Zhao, P.~Hu, L.~Nie, and X.~Zhao, ``{RoboChat}: A unified {LLM}-based interactive framework for robotic systems,'' in \emph{Int. Conf. on Rob., Intel. Cont. and Art. Intel.}, 2023, pp. 466--471.

\bibitem{wang2024prompt}
Y.-J. Wang, B.~Zhang, J.~Chen, and K.~Sreenath, ``Prompt a robot to walk with large language models,'' in \emph{Conf. on Dec. and Cont.}, 2024.

\bibitem{Kannan2024}
S.~S. Kannan, V.~L.~N. Venkatesh, and B.-C. Min, ``{SMART-LLM}: Smart multi-agent robot task planning using large language models,'' in \emph{IEEE/RSJ Int. Conf. on Intel. Rob. and Sys.}, 2024, pp. 12\,140--12\,147.

\bibitem{Wang2025}
J.~Wang, E.~Shi, H.~Hu, C.~Ma, Y.~Liu, X.~Wang, Y.~Yao, X.~Liu, B.~Ge, and S.~Zhang, ``Large language models for robotics: Opportunities, challenges, and perspectives,'' \emph{Journal of Automation and Intelligence}, vol.~4, no.~1, pp. 52--64, 2025.

\bibitem{kollar2013}
T.~Kollar, S.~Tellex, M.~R. Walter, A.~Huang, A.~Bachrach, S.~Hemachandra, E.~Brunskill, A.~Banerjee, D.~Roy, S.~Teller \emph{et~al.}, ``Generalized grounding graphs: A probabilistic framework for understanding grounded language,'' \emph{Journal of Artificial Intelligence Research}, pp. 1--35, 2013.

\bibitem{jiang2023vima}
Y.~Jiang, A.~Gupta, Z.~Zhang, G.~Wang, Y.~Dou, Y.~Chen, L.~Fei-Fei, A.~Anandkumar, Y.~Zhu, and L.~Fan, ``{VIMA}: General robot manipulation with multimodal prompts,'' in \emph{International Conference on Machine Learning}, 2023.

\bibitem{Liu2024b}
H.~Liu, Y.~Zhu, K.~Kato, A.~Tsukahara, I.~Kondo, T.~Aoyama, and Y.~Hasegawa, ``Enhancing the {LLM}-based robot manipulation through human-robot collaboration,'' \emph{IEEE Robotics and Automation Letters}, vol.~9, no.~8, pp. 6904--6911, 2024.

\bibitem{Boularias2015}
A.~Boularias, F.~Duvallet, J.~Oh, and A.~Stentz, ``Grounding spatial relations for outdoor robot navigation,'' in \emph{IEEE International Conference on Robotics and Automation}, 2015, pp. 1976--1982.

\bibitem{Wen2023}
S.~Wen, X.~Lv, F.~R. Yu, and S.~Gong, ``Vision-and-language navigation based on cross-modal feature fusion in indoor environment,'' \emph{IEEE Trans. on Cog. and Dev. Sys.}, vol.~15, no.~1, pp. 3--15, 2023.

\bibitem{hu2019}
Z.~Hu, J.~Pan, T.~Fan, R.~Yang, and D.~Manocha, ``Safe navigation with human instructions in complex scenes,'' \emph{IEEE Robotics and Automation Letters}, vol.~4, no.~2, pp. 753--760, 2019.

\bibitem{radford2022robust}
A.~Radford, J.~W. Kim, T.~Xu, G.~Brockman, C.~McLeavey, and I.~Sutskever, ``Robust speech recognition via large-scale weak supervision,'' \emph{arXiv 2212.04356}, vol.~10, 2022.

\bibitem{valin2018hybrid}
J.-M. Valin, ``A hybrid {DSP}/deep learning approach to real-time full-band speech enhancement,'' in \emph{IEEE Int. Workshop on Multimedia Signal Processing}, 2018, pp. 1--5.

\bibitem{ide1998introduction}
N.~Ide and J.~V{\'e}ronis, ``Introduction to the special issue on word sense disambiguation: the state of the art,'' \emph{Computational linguistics}, vol.~24, no.~1, pp. 1--40, 1998.

\bibitem{bubeck2023sparks}
S.~Bubeck, V.~Chadrasekaran, R.~Eldan, J.~Gehrke, E.~Horvitz, E.~Kamar, P.~Lee, Y.~T. Lee, Y.~Li, S.~Lundberg \emph{et~al.}, ``Sparks of artificial general intelligence: Early experiments with gpt-4,'' 2023.

\bibitem{yoo2021gpt3mix}
K.~M. Yoo, D.~Park, J.~Kang, S.-W. Lee, and W.~Park, ``{GPT3Mix}: Leveraging large-scale language models for text augmentation,'' \emph{arXiv 2104.08826}, 2021.

\bibitem{Acheampong2021}
F.~A. Acheampong, H.~Nunoo-Mensah, and W.~Chen, ``Transformer models for text-based emotion detection: a review of bert-based approaches,'' \emph{Art. Intel. Review}, vol.~54, no.~8, pp. 5789--5829, 2021.

\bibitem{Chi2023diffusion}
C.~Chi, Z.~Xu, S.~Feng, E.~Cousineau, Y.~Du, B.~Burchfiel, R.~Tedrake, and S.~Song, ``Diffusion policy: Visuomotor policy learning via action diffusion,'' \emph{The International Journal of Robotics Research}, p. 02783649241273668, 2023.

\bibitem{Baevski2020}
A.~Baevski, H.~Zhou, A.~Mohamed, and M.~Auli, ``wav2vec 2.0: a framework for self-supervised learning of speech representations,'' in \emph{Int. Conf. on Neural Information Processing Systems}, 2020.

\bibitem{Driess2023}
D.~Driess, F.~Xia, M.~S.~M. Sajjadi, C.~Lynch, A.~Chowdhery, B.~Ichter, A.~Wahid, J.~Tompson, Q.~Vuong, T.~Yu, W.~Huang, Y.~Chebotar, P.~Sermanet, D.~Duckworth, S.~Levine, V.~Vanhoucke, K.~Hausman, M.~Toussaint, K.~Greff, A.~Zeng, I.~Mordatch, and P.~Florence, ``{PaLM-E}: an embodied multimodal language model,'' in \emph{International Conference on Machine Learning}, 2023.

\bibitem{Raffel2020}
C.~Raffel, N.~Shazeer, A.~Roberts, K.~Lee, S.~Narang, M.~Matena, Y.~Zhou, W.~Li, and P.~J. Liu, ``Exploring the limits of transfer learning with a unified text-to-text transformer,'' \emph{J. Mach. Learn. Res.}, vol.~21, no.~1, Jan. 2020.

\bibitem{Zhou2019}
X.~Zhou, Y.~Gao, and L.~Guan, ``Towards goal-directed navigation through combining learning based global and local planners,'' \emph{Sensors}, vol.~19, 01 2019.

\end{thebibliography}
